# Digraph of Senegal's local languages: issues, challenges and prospects of their transliteration

## Authors' names

NGUER El hadji Mamadou , BAO Diop Sokhna, FALL Yacoub Ahmed, KHOULE Mouhamadou
Université Gaston Berger (UGB), Saint-Louis Sénégal
emnguer@ugb.edu.sn, baosokhna@hotmail.com, yacoubahmedfall@gmail.com, mouhamadoukhoule@gmail.com

**Abstract**

The local languages in Senegal, like those of West African countries in general, are written based on two alphabets: supplemented Arabic alphabet (called Ajami) and Latin alphabet. Each writing has its own applications. Ajami writing is generally used by people educated in Koranic schools for communication, business, literature (religious texts, poetry, etc.), traditional religious medicine, etc. Writing with Latin characters is used for localization of ICT (Web, dictionaries, Windows and Google tools translated in Wolof, etc.), the translation of legal texts (commercial code and constitution translated in Wolof) and religious ones (Quran and Bible in Wolof), book edition, etc. To facilitate both populations' general access to knowledge, it is useful to set up transliteration tools between these two scriptures. This work falls within the framework of the implementation of project for a collaborative online dictionary Wolof (Nguer E. M., Khoulé M, Thiam M. N., Mbaye B. T., Thiaré O., Cissé M. T., Mangeot M. 2014), which will involve people using Ajami writing. Our goal will consist, on the one hand in raising the issues related to the transliteration and the challenges that this will raise, and on the other one, presenting the perspectives.

**Keywords: Language, automatic processing of** language, transliteration, Ajami, supplemented Arabic characters, harmonized Quranic characters.

## 1. Introduction

According to the 2014 report (AUF report, 2014) of the International Organization of the Francophonie, the number of Francophones (actual and occasional) in Senegal is 4,277,000 out of a population of 14,133,280 in 2013 (World Bank Report, 2013), which means that 9,856,280 Senegalese do not understand the official language (French) for information and access to knowledge, training, legal texts, etc., guaranteeing a real and sustainable economic development.

Thus, French by itself fails to inform and train our people to get them out of ignorance and poverty. Consequently, the widespread use of local languages such as Wolof (vehicular language spoken by more than 80% of the population) remains the only alternative.

This language has long been written in completed non harmonized Arabic characters (commonly known as Wolofal) since the first contacts between the local population and the Arab-Muslim culture during the eighth and ninth century (CISSE M., 2006) . Today, more specifically since 1971, it is officially written with Latin letters (DECREE n° 2005-992, 2005), but also with harmonized Koranic characters (CCH) since 2007[1].

Despite the standardization and harmonization, non-harmonized writing continues to be dominant because of the lack of a state policy to popularize harmonized Koranic characters and the reluctance of people. It is commonly used among populations from Koranic schools (daaras) for communication, the management of current affairs (by a lot of traders, self-employed craft workers and farmers), literature (the religious texts, poetry, etc.), traditional religious medicine, etc.

It is important to note that the easiest way for these people to exploit their assets is to use Ajami. Arabic just constitutes a means to learn to read and write.

The writing with Latin characters is used for access to ICT (Web, dictionaries, tools for Windows and Google translated in Wolof, etc.), legal (commercial code, constitution translated in Wolof) and religious texts (Quran and Bible in Wolof), etc.

It is important to note that Ajami writing has always been marginalized since colonization. Today, despite the fact that in most West African countries, Senegal included, nearly 60% of the population has learned in his youth the Arabic script in the context of religious obligations, this writing continues to be marginalized. Indeed:

- 40% of the budget allocated to education goes exclusively to the French school.
- Official research, even today, on the local languages is performed based on the Latin alphabet.
- The publication of books in local languages, the translation of official documents and the location of ICT tools are based exclusively on this writing.
- Advertising signs, titles of TV shows are written in Latin characters

.

In short the fundamental mission of the Koranic school still remains to teach reading, writing and the memorization of the Koran and Muslim precepts

In sum, behind this digraph of local vehicular languages (Latin and Arabic characters) coexist two worlds that ignore each other. The official one uses Latin characters and draws its strength from official decrees. As for the other one, it only receives little institutional support and remains informal even though it is widespread and well integrated,.

[1] Officially launched in October 2007.

To help solve this problem and allow general access to knowledge (ICT, legal documents, religious texts, etc.) by these two groups of people, independently of writing, it is useful to set up automated tools transliteration of the two writings.
Our goal in this work is first to present the issues related to the transliteration and the challenges that it will bring about; then, a presentation of the project of automatic transliteration of local languages will follow.

Let us remind also that this work falls within the framework of the implementation of the project of a collaborative online dictionary for Wolof (Nguer E. M., Khoulé M, Thiam M. N., Mbaye B. T., Thiaré O., Cissé M. T., Mangeot M., 2014). Indeed, it aims at involving people using the Arabic alphabet supplemented in the project by showing them the dictionary based on Ajami writing through automatic transliteration. Such a dictionary could then be used in the daaras modernization policy.

In the rest of the document, we will:

- In the first part, present some transliteration concepts,
- Then address issues related to this transliteration in the second part,
- See later in the third part the challenges transliteration will raise,
- And finally, as a perspective, we will present in the fourth part our project to set up a tool of automatic transliteration of local languages.

## 2. Transliteration concepts

### 2.1. What is transliteration?

Transliteration is the operation that consists in replacing each grapheme of a writing system by a grapheme or group of graphemes from another system, regardless of pronunciation. A grapheme is the smallest contrastive unit in the writing system of a language[2]. The operation is reversible and therefore depends on the target writing system, but not the language. For example, Figure 1 shows the correlation table between the two alphabets for some grapheme of Senegal's local languages.

| ARABE | WOLOF | PULAAR | SONINKE | MANDINKA | SEREER | JOOLA | BALANT | SAAFI | LATIN | SenAjami2OT |
|---|---|---|---|---|---|---|---|---|---|---|
| x | x | x | x | x | x | x | x | x | b | ب |
|  | x | x | x | x | x | x |  |  | p | پ |
|  |  | x |  |  | x |  | x | x | ɓ | ٻ |
|  |  |  |  |  | x |  |  |  | ƥ | ب |
|  |  | x |  |  | x |  |  | x | ƴ | ج |

Fig. 1: Correspondence table for a few graphemes of Senegal's local languages.

| Wolof letter | Example in Wolof | Ajami letter | Signification in English | Example in Ajami |
|---|---|---|---|---|
| à | Jàng | أ | To learn | جٛنْگْ |
| c | car | ݖ | A branch | ݖَرْ |
| e | Ker | إ | A shadow | كࣹرْ |
| é | wér | إ | To be cured | وࣹرْ |
| ë | Bët | أ | An eye | بٛتْ |
| g | garab | گ | A drug | گَرَبْ |
| nt | bant | نت | A peace of wood | بَنتْ |
| ñ | ñaw | ݧ | To sew | ݧَوْ |
| ŋ | ŋaam | ڠ | A jaw | ڠَامْ |

Fig. 2: Correspondence table for a few graphemes of Wolof language

As shown in Figure 2, the use of diacritics or digraphs solves the problem of the unequal number of characters in the two alphabets writing systems.

This conversion process is primarily intended to enable the automatic and unambiguous reconstruction of the original writing (also known as retroconversion). In other words, the transliteration of a text must reproduce exactly the equivalent of the original text. To do so, ISO is used for standardization.

Figure 3 is an example of a transliteration of a text in Wolof.

[2] Crystal, David, 1997, The Cambridge Encyclopedia of Linguistics, second edition, UK: Cambridge University Press.

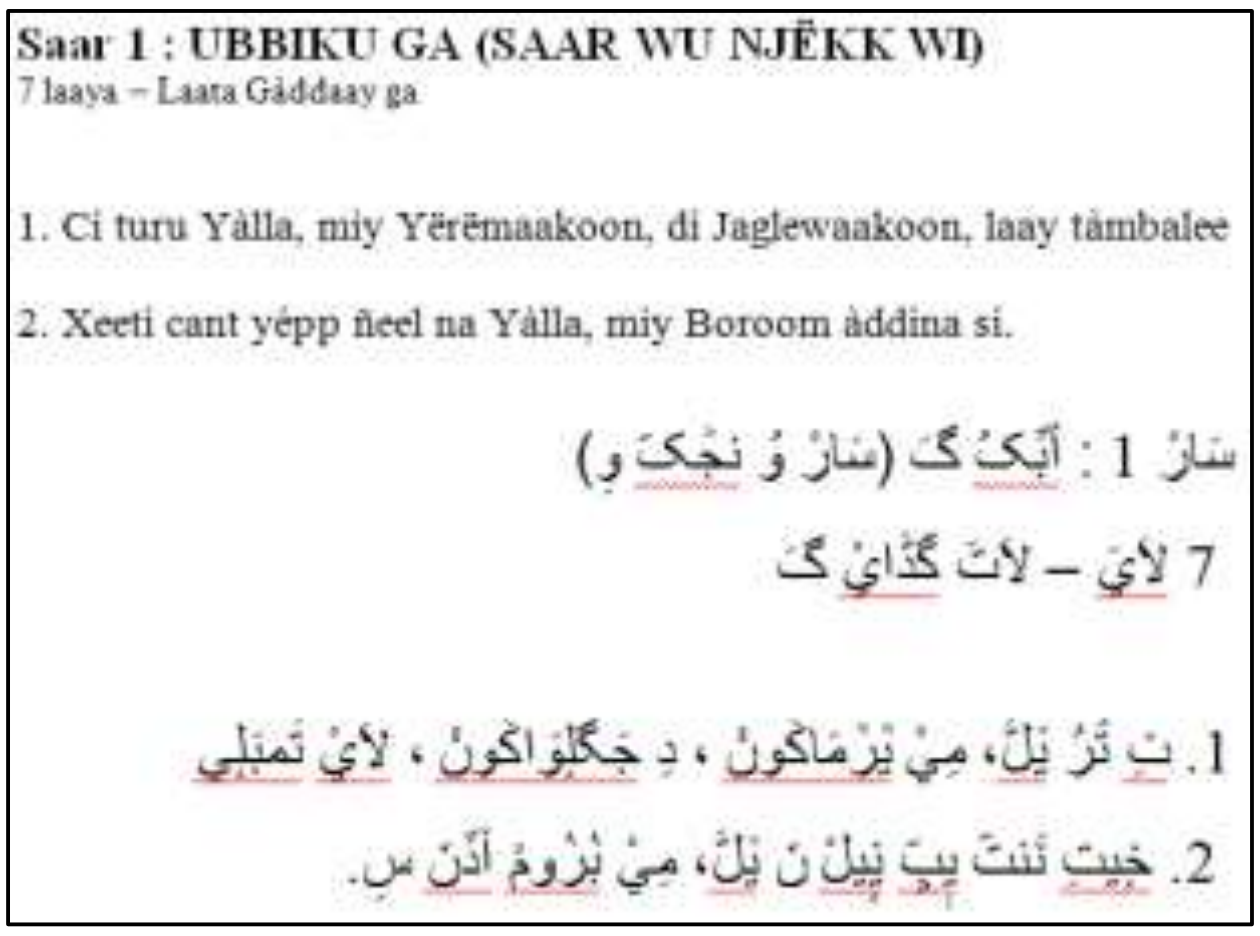

Fig. 3: Example of text in Wolof transliterated with macro (Paul-timothy, 2015).

Transliteration is not a translation. In fact, during the transliteration process, a word written in a writing system (alphabetical or syllabary[3]) is transposed to another; for example, from the Latin alphabet to Ajami Wolof alphabet. In other words, no translation is involved in this process. If the source word means nothing in the language in question, its transliteration does not mean more, even if it could give the impression of being a word in the language as it is written in its alphabetical or syllabary system.

## 2.2. What is the usefulness of transliteration?

It is mainly used by libraries in the bibliographic processing and the construction of indexes or for computer processing of textual data
Indeed, when a user performs searches and indexes contents, transliteration allows retrieving information written in a different alphabet and returns them to the user's writing system.

The transliteration also allows using a keyboard based on one alphabet to type a text into another alphabet. For example, it is possible with this technique to use a QWERTY keyboard to type text in Ajami.

In our situation, it will allow people to do Google searches with the writing of their choice (e.g. in Wolof Ajami or in Wolof Latin) and to find resources on the web written in Wolof (Latin or Ajami).

[3] Set of written symbols in which each symbol represents a phoneme but not a syllable. A syllable is an intermediate linguistic unity between the phoneme and word, characterized by the fusion of phonic units that constitute it. (We distinguish open syllable, ending in a sharp vowel, and the closed or obstructed syllable terminated by one or more consonants pronounced.)

# 3. Problematic of the transliteration of local languages

Several problems arise from the automatic transliteration of local languages. Among them, we can mention:

## 3.1. Problems related to keyboards

We cannot talk about transliteration without mentioning keyboard for text entry. There are two types of keyboards: physical and virtual keyboards. In general, a standardized physical keyboard designed for our endogenous languages does not yet exist.

Two solutions to this problem come out. The first one consists in reprogramming[4] standardized physical keyboards (e.g. reprograming the French one to write in Latin character or the Arabic one to write Ajami) so as to adapt them to our language through modifications of some keys. Although this solution is easy to implement (see Paul-timothy, 2015), it should be noted that the reallocation of the keys is generally not harmonized, and in addition, reassigned characters are often not engraved on the keys.

Therefore, if there are no standardized physical keyboards for Ajami and Latin alphabets for endogenous languages, one could put in place, for each endogenous language, a harmonized official method of adaptation of our keyboards.
The second solution would be to use virtual keyboards, the major drawback of which is the slowness in the entry of text.

## 3.2. Problems related to operating systems.

It is clear that our people cannot truly benefit from the advantages of ICT if the operating system they use utilizes a language they do not understand, French for example.. This is the main reason why Microsoft localized Windows 8 in Latin Wolof. The only thing left is now to present the Wolof Ajami system to enable people to use this script to enjoy the benefits of this tool.
Similar work should be done to localize in Wolof the Linux operating system which, unlike Windows, is free.

As smartphones and tablets are about to surpass computers in the field of access to ICT, it is also necessary to localize Android, the most used mobile operating system (85%), in Wolof.

## 3.3. Problems related to the popularization of writing Ajami

Whereas writing the West African languages with Ajami alphabet previously faced the absence of harmonization of its characters, its new standardized and harmonized version has only met limited success today as a mass literacy method. Several reasons explain this failure: the absence of popularization resulting from a lack of financial means and the distrust on the part of the populations who criticize governments for combating the traditional Koranic school and Islam above all.

[4] With software such as the Keyman program Tavultesoft change utility or the keyboard keys from Microsoft (MSKLC).

There's also the problem of segmenting words in Ajami writing. While establishing HQC (Harmonized Quranic characters), transcription standards based on Arabic characters were adopted with the elaboration of consensual rules of spelling and word segmentation.

Users do not know about the standardization of this form of writing, which allows us to state that there is a real problem of popularization.

To popularize effectively this form of writing, it is necessary:

- To organize meetings or seminars to involve all users and inform them of the existence of HQC.
- To create centers of experimentation in the regions emblematic of the use of Ajami to teach the people who use this form of writing to read and write using the standardized method.

## 4. The challenges of this transliteration

Among the challenges the automatic transliteration of our languages will face, we can mention:

### 4.1. The transliteration of text document

The establishment of transliteration macro in text editors like (Microsoft Word and OpenOffice) is of paramount importance especially as it will allow our writers to automatically produce two versions of each written document (one in Latin characters and another one in Ajami characters).

Such a macro (Paul-timothy, 2015) already exists in Word for Wolof and Sereer language but only transcribes Roman texts in Ajami writing; the reverse is not yet feasible.

Thus, there is a need to set up a macro to perform the automatic transliteration Ajami texts in Latin script.

Apart from text editors, it would also be nice to have applications on computer and mobile phone allowing the automatic transliteration of texts.

A first version of such an application was made in 2011 by the UGB's NLP team (9) for the Wolof language.

Although it allows transliteration in both ways, it does not support the harmonized Quranic characters. A new version is being developed to support harmonized Quranic characters. It will be followed by a generic version taking into account other local languages that will be selected.

In summary, only two alternatives exist for the transliteration of text document: a Word macro for the transliteration of Roman texts to Ajami writing and a Java application not taking into account the harmonized Quranic characters.

### 4.2. The transliteration of web page

Automatic transliteration of web pages will enable to bring both communities closer, at least as far as the access and sharing of resources and information across the Web are concerned. This will allow the Ajami community in particular to appropriate the benefits of the Web (E-commerce, E-banking, educational games, advertising of their works, etc.)

It is true that web page transliteration projects exist throughout the world for other languages (see UQAILAUT Project, 2012), but so far no known work was conducted for the transliteration of web pages for local languages in Senegal.

### 4.3. The transliteration of email

According to global statistics site in real time, planetoscope.com, 3.4 million emails are sent worldwide every second, or 107,000 billion annually, with 14,600 annual mails per person. This illustrates the important place of e-mail in the communication between people around the world.

Given this importance of emails, it is necessary to facilitate communication between the two communities. Consequently, it would be helpful to develop extensions for mail clients[5] like Mozilla Thunderbird and Microsoft Outlook and webmails (Gmail, Yahoo, Hotmail, etc.) to transliterate email.

This will contribute to significantly reduce the socio-economic barriers created by the digraph of the Senegal's local languages.

Again, such extensions exist for other languages (see UQAILAUT Project, 2012) but so far no known work was conducted for the transliteration of emails in Senegal's local languages.

### 4.4. The transliteration of SMS and MMS

Like email, SMS (Short Message Service) and MMS (Multimedia Messaging Service) represent large public media, quick and easy to adapt means of communication facilitating the reduction of socio-economic barriers created by the digraph of local languages.

It should be noted that none of our local languages is integrated for the moment in mobile operating systems (such as Android and Windows Phone), as is the case for Wolof in traditional operating systems (8 Windows, Linux, etc.). However, it is possible to develop for these systems applications to automatically transliterate SMS and MMS between the two writings.

---

[5] An email client is software that is used to read and send emails.

## 5. The prospects

In the framework of the project (Nguer E. M., Khoulé M, Thiam M. N., Mbaye B. T., Thiaré O., Cissé M. T., Mangeot M. , 2014) for the setting up of a standardized online collaborative dictionary for Wolof conforming with LMF standard (Francopoulo G., George M., Calzolari N, Monachini M., Bel N., Pet M., Soria C. , 2006), we intend to give the actors (linguists, lexicologists, lexicographers, writers, researchers, etc. ) the possibility of proposing new items, updating entries, and exporting the database for other purposes. And to benefit from the populations' expertise regardless of spelling, the website of the dictionary will be multilingual (Latin Wolof, French and English) and will include an automatic system of transliteration from Latin Wolof to Ajami Wolof. Furthermore, it will allow:

- The proposal of entries in both spellings.
- Discussion forums in both spellings
- To export the database in both spellings.
- Etc.

In all cases, the system will handle the need for automatic transliteration from Latin Wolof to Ajami

Also, in order to accompany people using Ajami in the production of documents in Latin characters, we will set up a macro (for Microsoft Word and OpenOffice) to make automatic transliteration of Ajami texts in Latin script going from the macro (Paul-timothy, 2015).

## 6. Conclusion

In Senegal, as in most West African countries, the digraph of local languages has created two worlds that ignore each other. As we have already mentioned, there is an official spelling in Latin characters that draws its strength from decrees. The other one, which uses the Arabic script, is widespread and well integrated; however, it receives only limited institutional support and remains informal. This work was carried out to help solve this problem through the automatic transliteration of the two writings,.

Its aim is, on the one hand, to present the issues related to the transliteration and, on the other, the challenges that it will raise.

Besides, it will help develop later a macro (for Microsoft Word and OpenOffice) to perform an automatic transliteration of Ajami texts in Latin script (Paul-timothy, 2015).
Also, in the framework of the project (Nguer E. M., Khoulé M, Thiam M. N., Mbaye B. T., Thiaré O., Cissé M. T., Mangeot M. , 2014), it will set up an automatic system of transliteration from the Latin Wolof to the Ajami Wolof to benefit from the expertise of populations without distinction of spelling. In addition it will propose inputs in both spellings to post topics in discussion forums and export the database for other purposes.

The final goal is to provide general access to knowledge (ICT, legal, religious texts, etc.) to these two groups regardless of the spelling used. This will help significantly reduce the socio-economic barriers created by the digraph of local languages.